\documentclass[sigconf]{acmart}
\usepackage{bbding}
\usepackage{pifont}
\usepackage{amsmath,amsfonts}
\usepackage{bbm}
\usepackage{algorithm}
\usepackage{algorithmic}
\usepackage{graphicx}
\usepackage{slashbox}
\usepackage{textcomp}
\usepackage{xcolor}
\usepackage{makecell}
\usepackage{algorithm}
\usepackage{algorithmic}
\usepackage{subfig}
\usepackage{multirow}

\def\BibTeX{{\rm B\kern-.05em{\sc i\kern-.025em b}\kern-.08em
    T\kern-.1667em\lower.7ex\hbox{E}\kern-.125emX}}
\usepackage{stfloats}
\usepackage{hyperref}
\usepackage{cleveref}
\usepackage{ragged2e}
\crefname{figure}{Fig.}{Figs.}
\crefname{table}{Table}{Tables}
\crefname{algorithm}{Algorithm}{Algorithms}
\crefname{equation}{Eq.}{Eqs.}
\crefname{section}{Section}{Sections}

\newtheorem{definition}{Definition}
\AtBeginDocument{%
  \providecommand\BibTeX{{%
    \normalfont B\kern-0.5em{\scshape i\kern-0.25em b}\kern-0.8em\TeX}}}

\setcopyright{acmcopyright}
\copyrightyear{2022}
\acmYear{2022}
\acmDOI{xx.xxxx/xxxxxxx.xxxxxxx}

\acmPrice{15.00}
\acmISBN{978-1-4503-XXXX-X/18/06}

\begin{document}

\title{PGX: A Multi-level GNN Explanation Framework Based on Separate Knowledge Distillation Processes}

\author{Tien-Cuong Bui}
\email{cuongbt91@snu.ac.kr}
\affiliation{%
  \institution{Dept. of ECE}
  \institution{Seoul National University}
  \country{South Korea}
}

\author{Wen-syan Li}
\email{wensyanli@snu.ac.kr}
\affiliation{%
  \institution{Graduate School of Data Science}
  \institution{Seoul National University}
  \country{South Korea}
}

\author{Sang-Kyun Cha}
\email{chask@snu.ac.kr}
\affiliation{%
  \institution{Graduate School of Data Science}
  \institution{Seoul National University}
  \country{South Korea}
}


\begin{abstract}
Graph Neural Networks (GNNs) are widely adopted in advanced AI systems due to their capability of representation learning on graph data. Even though GNN explanation is crucial to increase user trust in the systems, it is challenging due to the complexity of GNN execution. Lately, many works have been proposed to address some of the issues in GNN explanation. However, they lack generalization capability or suffer from computational burden when the size of graphs is enormous. To address these challenges, we propose a multi-level GNN explanation framework based on an observation that GNN is a multimodal learning process of multiple components in graph data. The complexity of the original problem is relaxed by breaking into multiple sub-parts represented as a hierarchical structure. The top-level explanation aims at specifying the contribution of each component to the model execution and predictions, while fine-grained levels focus on feature attribution and graph structure attribution analysis based on knowledge distillation. Student models are trained in standalone modes and are responsible for capturing different teacher behaviors, later used for particular component interpretation. Besides, we also aim for personalized explanations as the framework can generate different results based on user preferences. Finally, extensive experiments demonstrate the effectiveness and fidelity of our proposed approach.
\end{abstract}


\maketitle

\section{Introduction}
Graph Neural Networks (GNNs) \cite{wu2020comprehensive} are essential components in numerous real-world applications nowadays, such as recommendation systems \cite{ying2018graph, wang2019neural, wu2020graph}, social networks \cite{perozzi2014deepwalk, fan2020graph}, e-commerce websites \cite{wu2019session, zhou2021temporal}, and fraud detection \cite {li2019spam, dou2020enhancing, cheng2020graph}. As GNNs become more and more ubiquitous, we are interested in understanding their internal logic and influential factors in predictions. For instance, if a recommendation system suggests a user a movie, they want to know the reason for this suggestion. Besides, interpretation methods can also enhance trust and fairness in the system. It could be big trouble if suggestions are offensive due to gender discrimination or other social biases. Therefore, GNN explanation is mandatory to increase user trust in such systems.

GNNs suffer from the lack of transparency similar to other deep learning (DL) methods, thus making predictions unintelligible. Even worse, explaining GNN models is much harder than other DL models due to the complex message-passing patterns during execution. GNN predictions are impacted by node features, edge information, the graph structure, and observed node outcomes \cite{sen2008collective, wang2021combining}. While several GNN explanation methods have been proposed lately, they mainly focus on finding essential subgraphs via edge selection processes \cite{ying2019gnnexplainer, luo2020parameterized, schlichtkrull2020interpreting, yuan2020xgnn} or Monte Carlo tree search \cite{yuan2021explainability}. However, we argue that they lack generalization and suffer from a computational burden, especially when graphs are massive. Besides, each user has different preferences and expectations for explanations matching their prior belief \cite{miller2019explanation}. Therefore, explanations require multi-level and multi-granular outputs. Even though \cite{wang2021towards} proposes a multi-grained explanation method, it simply consists of two steps of edge selection and does not meet the requirements.

In this paper, we propose a multi-level GNN explanation framework to address the problems above, named PGX. From the data perspective, graph data consists of various data structures such as tabular data like node/edge features and the graph structure. Therefore, GNNs can be regarded as a multimodal learning process of multiple components in graph data. With that in mind, we break down the GNN explanation problem into simpler sub-problems categorized into a hierarchical structure presented in \cref{fig:granular}. Each level in our framework provides beneficial insights in a different way. Specifically, the highest level explanation measures the overall contribution of individual components by answering questions like "What are attributions of graph components to the model execution and predictions?". We adopt the non-additive interaction set concept proposed in \cite{tsang2020does} to answer this type of question. Furthermore, we relax the complexity of the problem by considering only node features and the graph structure in the interaction set, which is equivalent to measuring each component's marginal contribution. This relaxation also enables PGX to deliver explanations at fine-grained levels.

Next, we introduce a class of methods based on knowledge distillation \cite{hinton2015distilling} to measure particular feature attributes and the graph structure at fine-grained levels. Specifically, the second level of explanation is to obtain a general understanding of influential nodes in the graph and crucial attributes in node features, while the third level provides local explanations of predictions with specific neighboring impacts and feature attributions. Knowledge from the original GNN model is transferred to subsequent students, wherein each one is responsible for imitating a specific function of the teacher. For instance, a special graph-based student model tries to weight the interaction among nodes, while a non-graph-based neural network student is to capture the feature transformation process. Later, we implement the Personalized PageRank algorithm based on edge weights of the graph-based student for link analysis to find out globally important nodes and locally impactful neighbors. Furthermore, users can obtain interpretations of the graph structure via our interactive visualization. Next, the non-graph-based neural network is the input of feature attribution measurement methods \cite{lundberg2017unified, ribeiro2016should}. Each knowledge distillation process is a standalone procedure and executed only once, corresponding to each component. With the flexibility of student models, our approach is very promising and extensible. Furthermore, it also enables the explainer to provide customized explanations given different preferences towards personalization.

The main contributions of this paper are three-fold:
\begin{itemize}
\item We propose a multi-level GNN explanation framework, which provides a complete view of GNN explanations. The framework can deliver explanations with appropriate detailed information depending on specific requirements. 
\item We also propose leveraging knowledge distillation to transform original GNN models into simpler student models. By doing that, we can exploit existing explainable AI tools and link analysis algorithms for GNN explanations. 
\item We present extensive experiments with convincing evidence to verify the fidelity and consistency of proposed methods. 
\end{itemize}

The remainder of this paper is structured as follows. In \cref{related_work}, we introduce the related work. In \cref{preliminary}, we describe background concepts and problem formulation. Then we first describe the framework overview in \cref{method}. \cref{attribution_analysis} elaborates on attribution analysis processes based on knowledge distillation. In \cref{experiments}, experiments are presented. \cref{discussion} presents discussions with learned lessons and interesting observations. Finally, we conclude our work in \cref{conclusion_part}.

\section{Related Work}\label{related_work}

\subsection{Perturbation-based Methods}
Following GNNExplainer \cite{ying2019gnnexplainer}, multiple perturbation methods \cite{luo2020parameterized, schlichtkrull2020interpreting, yuan2021explainability} have been proposed to explain GNNs by extracting essential subgraphs. These methods focus on designing mask learning processes that act as an edge filter. 
However, as they all followed experimental settings of \cite{ying2019gnnexplainer}, which were executed on trivial synthetic datasets using a transductive style, it is difficult for these methods to operate on real-world problems when unknown nodes appear.
Besides, the datasets used for the node classification were not general, as graphs in many domains assume that nodes in a community should have the same label \cite{zhur2002learning}.  
Therefore, we argue that explanations of these methods are only appropriate for the graph classification problem as masks act as a global filter rather than paying attention to local impacts. Furthermore, the number of learned parameters linearly increases, corresponding to the number of edges, thus making it hard to scale. 


\subsection{Game-theoretic Approach}
Shapley value is a concept in game theory that distributes gains and costs to players working in a coalition, which was first applied to machine learning model explanations by \cite{lundberg2017unified}. In GNN prediction interpretation, SubgraphX \cite{yuan2021explainability} is the first work that utilizes the Shapley value. Specifically, it explains predictions by exploring subgraphs with a Monte Carlo tree search guided by the marginal contribution values. GraphSVX \cite{duval2021graphsvx} applied \cite{lundberg2017unified} to measure impact factors to predictions, wherein it assumed that nodes and features constitute a set of independent players. This assumption is unrealistic as the GNN aggregates neighbors' information multiple times before making predictions. Besides, as the number of players is enormous, the contribution is scattered, leading to unhelpful explanations. Therefore, we must carefully specify the set of players in a cooperative game to focus on crucial things only. This specification is beneficial to users as they have different concerns when looking at a prediction.

\subsection{Model-level Explanations}

Unlike instance-level methods, model-level methods provide high-level insights and understandings of model behaviors. Applying techniques like \cite{FeatureV47:online}  to explain GNN behaviors is troublesome as message-passing patterns are complicated. \cite{yuan2020xgnn} was the first work that tackled this problem via graph generation. Specifically, it trains a graph generator via policy gradient so that the generated graph patterns maximize a certain prediction of the model. However, the generation process also focuses only on finding globally important subgraphs making it less sensitive to the local structure of a certain prediction. 

Lately, Wang et al., 2021 \cite{wang2021towards} proposed a multi-grained approach for global and local explanations. First, it specifies the class-wise knowledge across instances with the same prediction via a class-aware attribution module, similar to \cite{ying2019gnnexplainer, luo2020parameterized}. As the coarse-grained edge selection fails to detect dependencies within the selected edges, an additional fine-tuning step optimizes the edge sampling process to generate an appropriate explanation for a certain prediction. However, it does not consider feature importances and impacts of neighboring labels. 

\subsection{Knowledge Distillation for GNNs}
Knowledge distillation \cite{gou2021knowledge} can be used to explain the overall logic behind the black box models \cite{alharbi2021learning} by making the student model transparent. 
Lately, it has been applied to GNNs, wherein student models try to reach the accuracy of teachers while requiring fewer computations. Xiang et al., 2021 \cite{deng2021graph} proposed distilling knowledge from pre-trained GNNs to a student model using generated fake graphs. Similarly, Chaitanya et al., 2021 \cite{joshi2021representation} investigated representation distillation by introducing a specialized loss function. Lately, Yuan et al., 2022 \cite{li2022egnn} applied \cite{hinton2015distilling} to construct an explainable GNN, which decomposes the interpretation process into multiple edge selection processes via attention mechanisms. Even though it provides a high-level understanding of model predictions, it encounters two fatal problems: Computation costs due to multi-hop decomposition; Trustability of the attention networks, which do not imitate message-passing patterns.

\begin{table}[htbp]
    \centering
    \resizebox{\columnwidth}{!}{%
    \begin{tabular}{r|l}
         \hline
         \textbf{Symbol} & \textbf{Definition}  \\
         \hline
         $\mathcal{G; V; E; X}$ & Graph; Node set; Edge set; Node features \\
         $f;g$               & A trained GNN model; A surrogate model \\
         $\mathcal{N}_i$        & A set of 1-hop neighbors of node $i$\\
         $m_{ij}$               & Message on the edge between node i and j\\
         $m^l_i$                & Aggregated message of node $i$ at layer $l$ \\
         $h_i^l;h^L$           & Node $i$'s embedding at $l^{th}$ and the last layer $L$\\
         $\mathcal{I} $         & An interaction set in a multi-modal approach \\
         $\mathcal{L}_{CE};\mathcal{L}_{DL};\mathcal{L}$ & Cross-entropy loss; Distillation loss; Overall loss \\
         $\lambda$              & A hyper-parameter to adjust $\mathcal{L}_{DL}$ \\
         $\omega$               & A function operates on elements in $\mathcal{I}$ \\
         $\mathcal{A}^* $       & A row-normalized matrix measuring node interactions \\
         $\mathcal{R_V};\pi $  & The rank vector of nodes; Preference vector\\
         $z_t;z_s;\tau$         & Teacher outputs; Student outputs; Temperature term \\
         $y;\hat{y};p$          & Label values; Predicted values; Softmax function \\
         \hline
    \end{tabular}
    }
    \caption{Glossary of Notations.}
    \label{tab:symbol_list}
\end{table}

\section{Preliminaries} \label{preliminary}
\subsection{Graph Neural Networks}
GNNs adopt the message-passing paradigm to propagate and aggregate information along edges in the graph. Specifically, each GNN layer involves three crucial steps: propagation, aggregation, and updating. First, a message $m^l_{ij} = \textrm{Message}(h^{l-1}_i, h^{l-1}_j)$, where $h^{l-1}$ is the representation of a node $v$ at layer $l-1$, is propagated to adjacent nodes. Second, for each node $v_i$, GNN aggregates messages received from its neighbors $\mathcal{N}_i$, wherein the aggregation operation is $m^l_i = \textrm{Aggregation}({m^l_{ij}|j \in \mathcal{N}_i})$. Lastly, given $m^l_{i}$ and $h^l_i$, GNN computes a new representation vector for $v_i$ at layer $l$ as follows: $h^l_i = \textrm{Update}(m^l_i, h^{l-1}_i)$. The representation vectors of the last layer $h^L$ are then used for downstream tasks, such as node classification.

\subsection{Knowledge Distillation}
Knowledge Distillation, first proposed in \cite{hinton2015distilling}, is originally used for model compression, wherein a small model (student) tries to mimic the large model (teacher)'s decisions. Gou et al., 2021 \cite{gou2021knowledge} presented a complete survey of knowledge distillation, including various techniques and approaches. Here, let us briefly summarize the original method \cite{hinton2015distilling}.

\begin{equation}
    p(z_i,\tau) = \frac{exp(z_i/\tau)}{\sum_j exp(z_j/\tau})
    \label{softmax_tau}
\end{equation}

\begin{figure}[htbp]
    \centering
    \includegraphics[width=\columnwidth, trim={3.5cm 3.5cm 4.2cm 3.7cm}, clip]{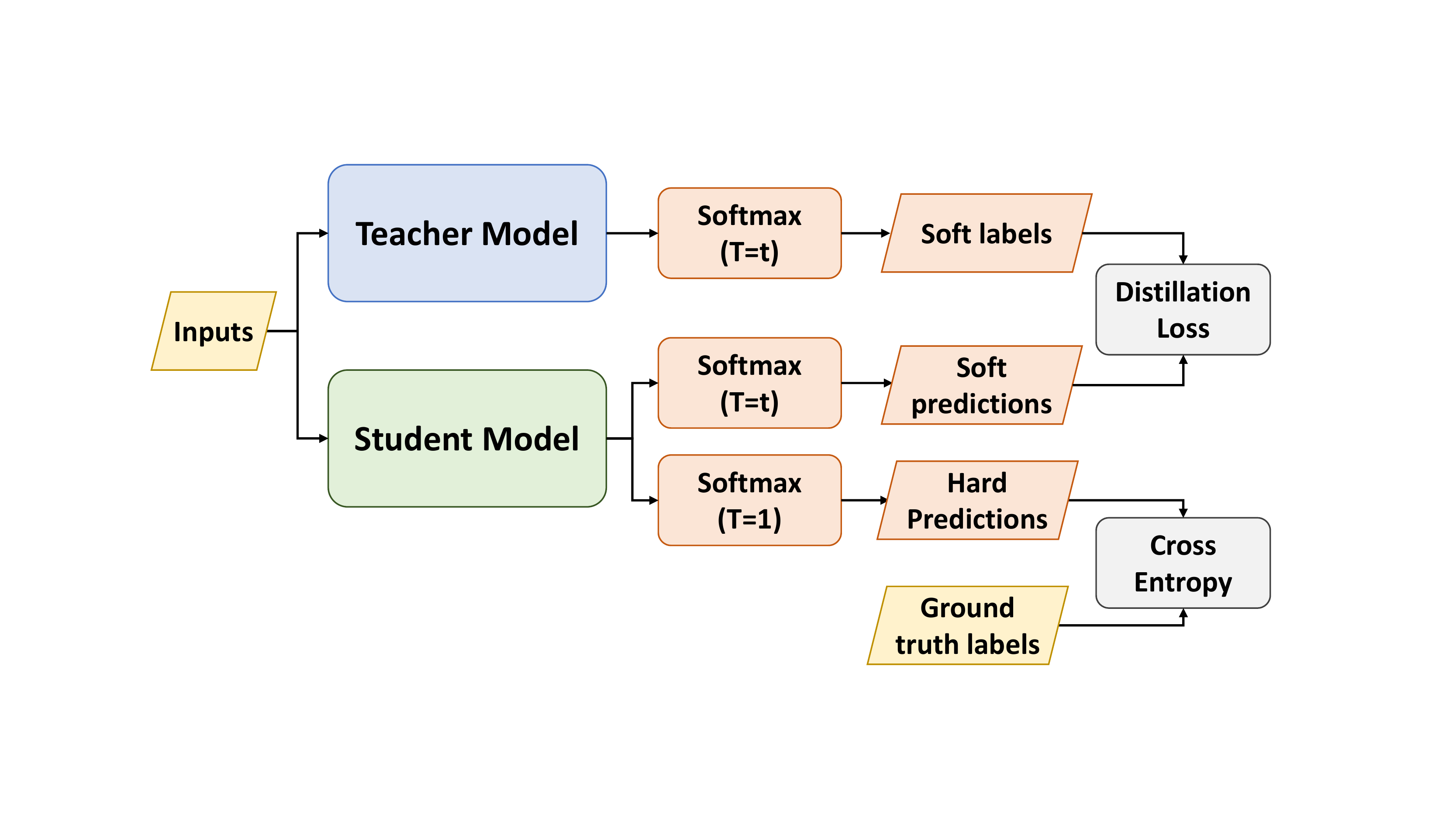}
    \caption{Knowledge Distillation Paradigm \cite{hinton2015distilling}. The student's loss is regularized by the knowledge distillation loss, which tries to bring the student's predictive distributions close to the teacher's.}
    \label{fig:knowledge_distillation}
\end{figure}

In \cref{fig:knowledge_distillation}, given the same set of input features, both models first output so-called soft labels/predictions, which are results of softmax function with a temperature term $\tau$ shown in \cref{softmax_tau}. Then, the student model computes the distillation loss, usually a soft cross-entropy loss function. After that, it also generates hard predictions to compare with the ground-truth labels via the cross-entropy loss function. Finally, we have the student loss function as follows:

\begin{equation}
    \mathcal{L} = \mathcal{L}_{CE}(y, p(z_s,1)) + \lambda \mathcal{L}_{DL}(p(z_t, \tau), p(z_s, \tau)),
\end{equation}
where $z_t, z_s$, and $y$ are teacher/student outputs and labels, respectively. $\lambda$ is a hyper-parameter that controls the magnitude of $\mathcal{L}_{DL}$.

\subsection{GNN Explanation: Problem Formulation}
Given a graph $\mathcal{G = (V,E,X)}$, a GNN model $f$ is trained to predict node/edge/graph level outcomes in classification problems. Explanations for GNN predictions aim at finding crucial factors contributing the most to the model execution and prediction scores. Specifically, we construct a hierarchical structure of these factors, as presented in \cref{fig:granular}. This paper limits the scope to only node classification problems and discards edge features in all analyses for simplicity. We can easily extend the proposed solutions to other GNN problems but leave them for future work. 

\section{PGX: A Multi-level Explanation Framework} \label{method}
Many factors contribute to GNN execution and predictions as GNN models usually adopt message-passing paradigms. Node information is passed to neighbors in complicated patterns making the model hard to disentangle and explain. As opposed to existing approaches which only focus on specific aspects of explanations, we propose a unified framework to explain GNNs on multiple levels.

\begin{figure}
    \centering
    \includegraphics[width=\columnwidth]{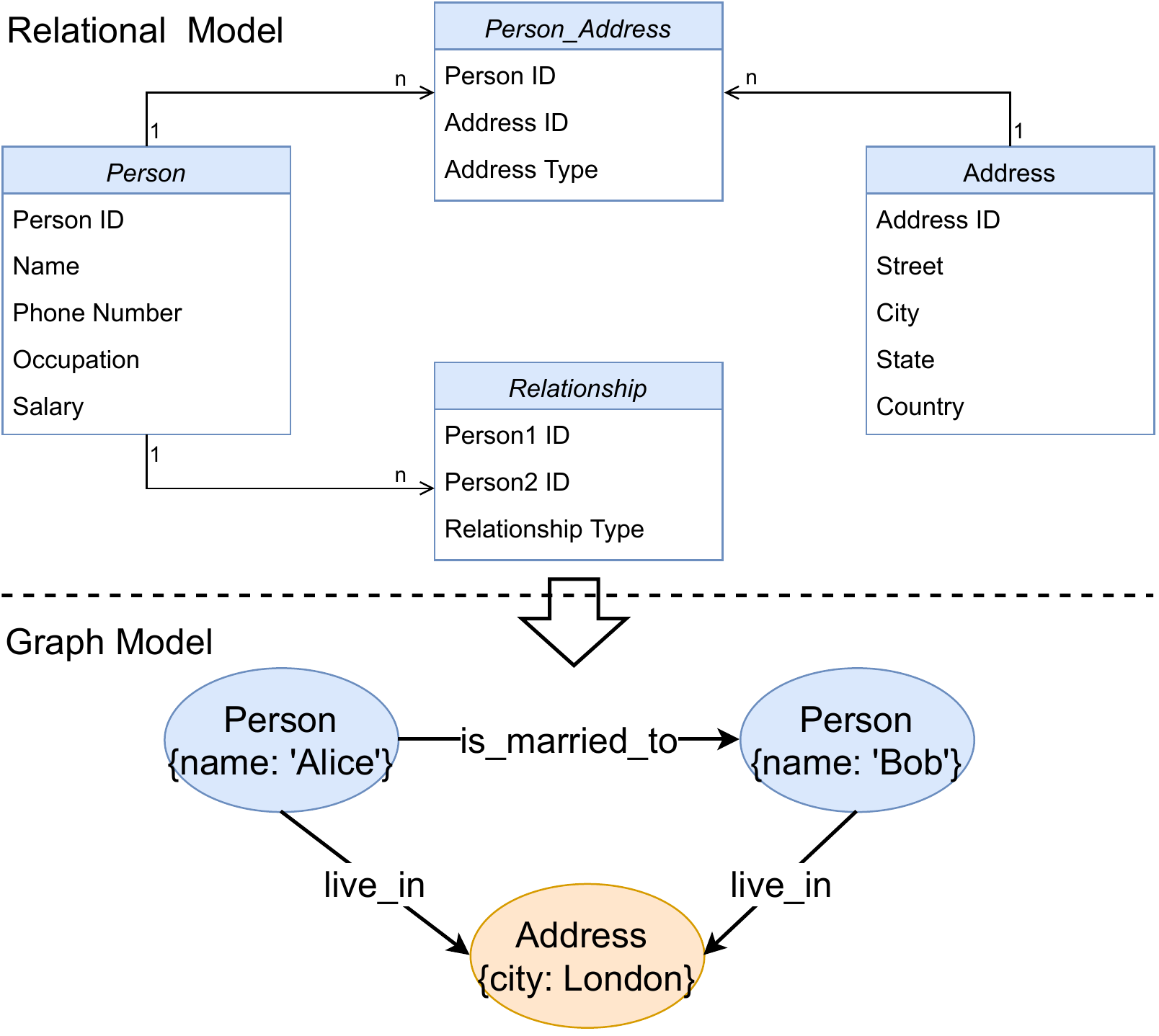}
    \caption{An Example of Relational Model to Graph Model.}
    \label{fig:rel_graph}
\end{figure}

\subsection{Framework Overview}

\begin{figure}[htbp]
    \centering
    \includegraphics[width=\columnwidth,trim={4.1cm 1cm 5.75cm 0.1cm},clip]{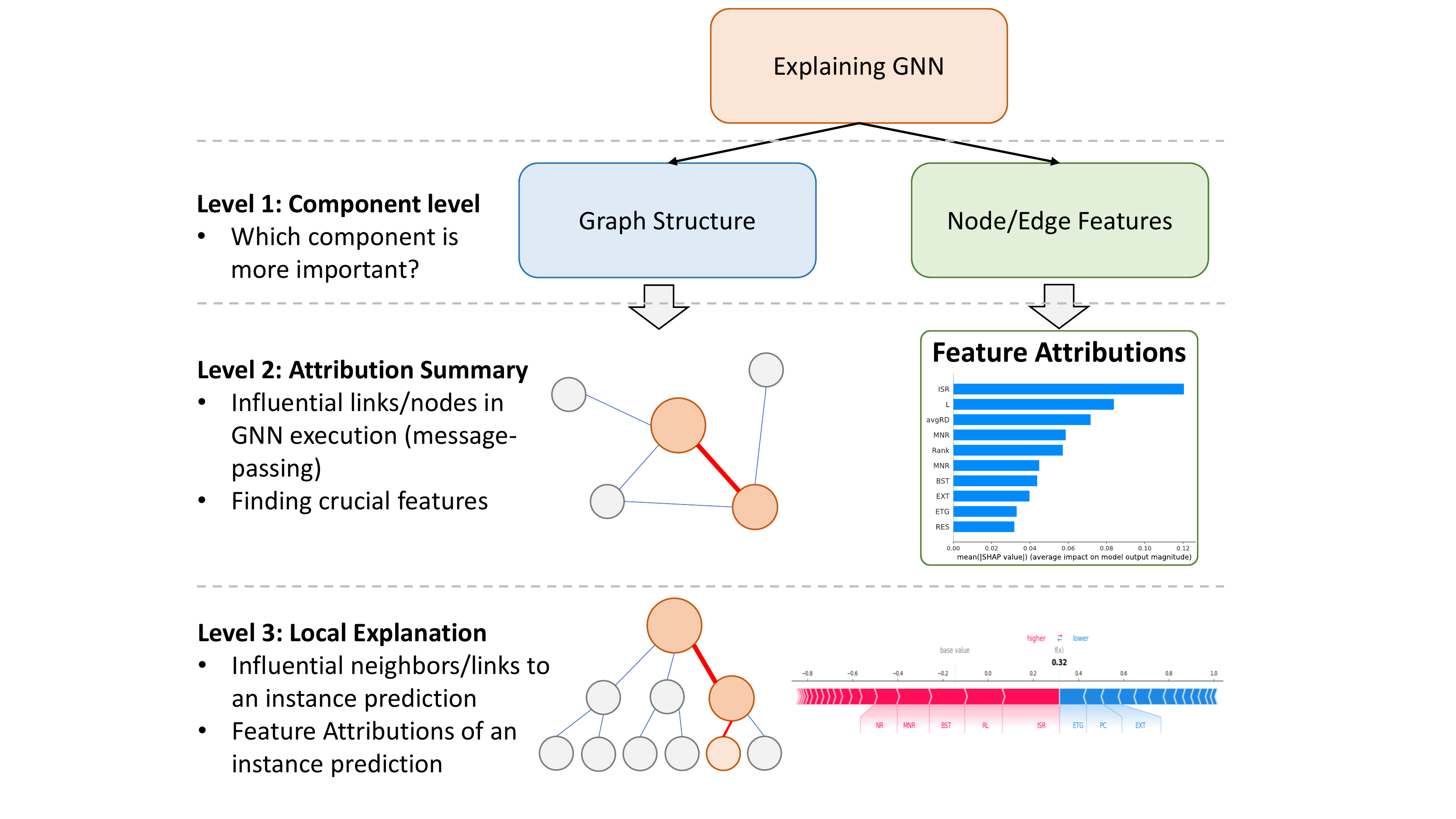}
    \caption{Multi-level GNN Explanations.}
    \label{fig:granular}
\end{figure}

\begin{definition}
A GNN model operates on a graph $G$ is a multimodal learning process that simultaneously handles different entities and relationships.
\end{definition}

A GNN model can be considered a multimodal approach if we look at it from the data point of view, as \cref{fig:rel_graph} shows. Specifically, graph data is the constitution of multiple entities (nodes), relationships (edges), and attributes. GNN models are efficient methods to model these factors simultaneously. Due to multimodal properties, providing a single explanation for GNNs is incomplete. Besides, from a user perspective, each person prefers different explanations for the same input graph and trained model. To solve these problems, we introduce a novel multi-level GNN explanation framework represented as a hierarchical structure in \cref{fig:granular}.  

Specifically, the framework provides explanations on multiple levels, from a global view to fine-grained attributions. At the top level, it measures the global attribution of each component in the graph data, such as the structure or features, by adopting the concept proposed in \cite{tsang2020does}. The second level focuses on essential factors within each particular component. In other words, it answers questions like "Which are the top influential nodes?" or "Which are the top influential features?". Next, the framework explains individual predictions by outputting neighbors' influences and local feature attributions in the last level. Besides, the last-level explanation is flexible as we can view local influences in multiple hops via an interactive visualization tool. To handle the last two levels, we use knowledge distillation to transfer knowledge from the base model to simpler ones. Particularly, each student has a role in imitating a specific function of the teacher. For instance, a special graph-based model captures the interactions among nodes, while a non-graph-based neural network model is trained to imitate the whole feature transformation process. After that, each one is responsible for the interpretation of a particular component. Please note that training one student does not affect the learning process of other students. Therefore, it can be done sequentially or simultaneously, depending on particular situations.

\subsection{Component-level Attribution Measurement} \label{component_level}
As the graph structure and node features are the main ingredients of GNN execution, ones often assume that they are equally important \citep{huang2020graphlime, duval2021graphsvx}. We argue that interactions in GNNs are non-additive attributions due to complex patterns of message-passing schemes. Following \cite{tsang2020does, tsang2017detecting, jin2019towards}, we define the component-level attributions as a non-additive interaction of nodes, edges, and features.

\begin{definition}
A function $f$ operating on $\mathcal{G = (V,E,X)}$ consists of a statistical non-additive interaction of multiple factors indexed in the set $\mathcal{I} \iff$ $f$ cannot be decomposed into a sum of $|\mathcal{I}|$ subfunctions $f_i$, which does not include $i^{th}$ interaction, such that: $f(x) \neq \sum_{i\in\mathcal{I}} f_i(x_{\backslash \{i\} })$.
\end{definition}

Let us assume that $f$ operating on $\mathcal{G}$ can be represented as a generalized additive function \cite{tsang2020does, tsang2017detecting} as follows:

\begin{equation}
    f(\mathcal{G}) = \sum_{i \in \mathcal{I}} \omega_i(x_{\mathcal{I}_i}) + b,
    \label{addition_g}
\end{equation}
where $x_{\mathcal{I}_i}$ corresponds to the element $i^{th}$ in $\mathcal{I}$, $\omega_i$ is a function of $x_{\mathcal{I}_i}$, and $b$ is regarded as a bias. Following \cite{tsang2020does}, we propose using reference values to calculate the attribution of each element in $\mathcal{I}$. Let us denote the reference value of $x_{\mathcal{I}_i}$ as $x'_{\mathcal{I}_i}$. The attribution of $\mathcal{I}_i$ is calculated as follows:

\begin{equation}
    \phi(\mathcal{I}_i) = \omega_i(x_{\mathcal{I}_i}) - \omega_i(x'_{\mathcal{I}_i}).
\end{equation}

\noindent\textit{Proof.} Based on \cref{addition_g}, we have:
\begin{equation}
\begin{aligned}
    \phi(\mathcal{I}_i) &= f(\mathcal{G}) - f(\mathcal{G}_{\backslash \{i\}}) \\
                        &= \left( \omega_i(x_{\mathcal{I}_i}) + \sum_{j \in \mathcal{I},i\neq j} \omega_j(x_{\mathcal{I}_j}) + b \right) \\
                        &\hspace{0.5cm} - \left( \omega_i(x'_{\mathcal{I}_i}) + \sum_{j \in \mathcal{I},i\neq j} \omega_j(x_{\mathcal{I}_j}) + b \right) \\
                        &= \omega_i(x_{\mathcal{I}_i}) - \omega_i(x'_{\mathcal{I}_{i}})
    \label{addition_proof}
\end{aligned}
\end{equation}

To relax the problem complexity, we limit the set of interactive factors to $\mathcal{I = \{E,X\}}$. In other words, we consider the graph structure and the set of node features as two interactions in $\mathcal{I}$. Therefore, we compute the marginal contribution (MC) for $\mathcal{I}_i$, as follows: 
\begin{equation}
    \begin{aligned}
    \phi(\mathcal{X}) &= f(\mathcal{G}) - f(\mathcal{G}_{\backslash \mathcal{X}}), \\
    \phi(\mathcal{E}) &= f(\mathcal{G}) - f(\mathcal{G}_{\backslash \mathcal{E}}).
    \end{aligned}
    \label{structure_cont}
\end{equation}

Here, $f$ outputs a predicted probability score for a given graph. In practice, reference values for node features are either fixed-value vectors or average-value vectors. However, generating a reference graph is difficult due to the network structure complexity. Inspired by perturbation techniques applied to images \cite{ribeiro2016should,lundberg2017unified}, we create a graph that contains only self-loop edges for the structure reference.   

\section{Fine-grained Attribution Analysis}\label{attribution_analysis}

Fine-grained attribution analysis corresponds to the bottom-two levels in \cref{fig:granular}, which is responsible for evaluating specific contributions of elements in the graph structure and node features. In \cref{component_level}, we assume that interactions only occur within an individual component, such as edges in the graph structure and attributes in node features. Based on this assumption, we evaluate fine-grained attribution for components separately using two different knowledge distillation procedures.

\begin{definition}
Given a GNN model $f$, fine-grained attributions can be obtained via separate knowledge distillation processes.
\end{definition}

\subsection{Node Feature Attribution}
\begin{figure}[htbp]
    \centering
    \includegraphics[width=\columnwidth,trim={0cm 3.5cm 2.5cm 6cm}, clip]{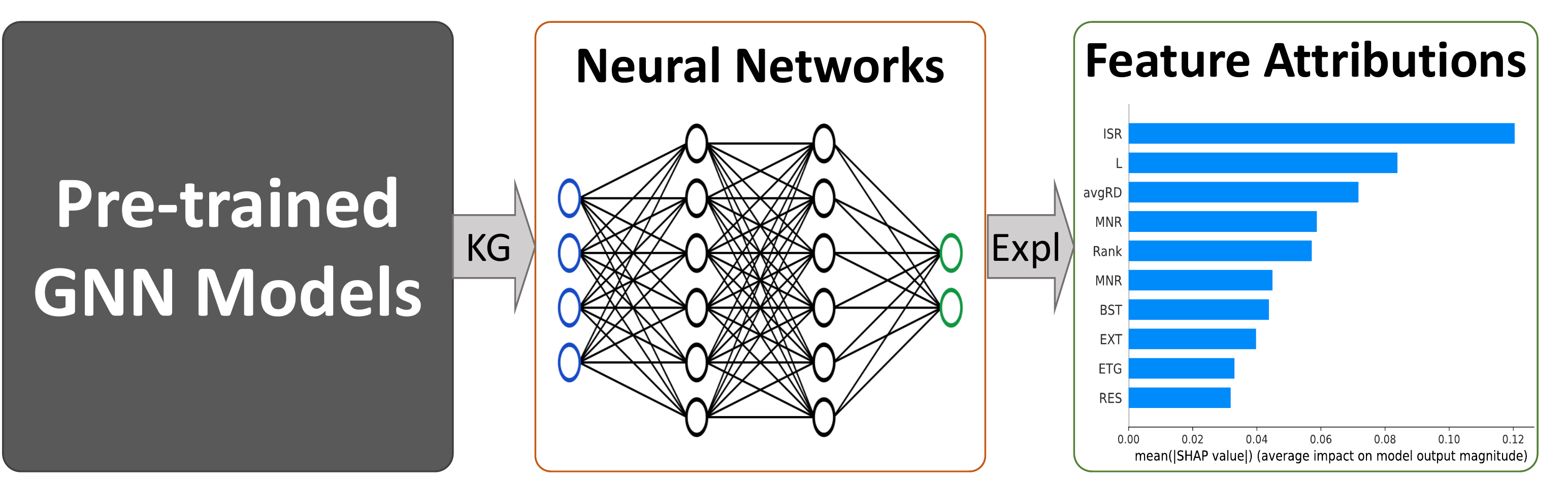}
    \caption{Execution Pipeline for Measuring Feature Attributions based on Knowledge Distillation. Pre-trained GNN models are distilled to non-graph-based neural networks used for feature attribution analysis.}
    \label{fig:feature_attribution}
\end{figure}

Feature attribution analysis is an active area in interpretable machine learning with numerous methods \cite{lundberg2017unified, ribeiro2016should, tsang2020does}. However, existing methods are inappropriate for GNNs due to complex message-passing patterns. We solve this problem indirectly using the knowledge distillation method, presented in \cref{fig:knowledge_distillation}. Specifically, a non-graph-based model is trained to capture the feature transformation process of the original GNN. As depicted in \cref{fig:feature_attribution}, any neural network that takes a set of samples as inputs can be a candidate for the student model. Practically, the student model is exposed to more labels, thus making it capable of imitating the GNN teacher behaviors. After that, the student model is plugged into feature attribution methods to obtain a global summary of feature importances or a local explanation of an instance. As feature attribution methods have been continuously proposed, our approach enables one to integrate these methods into GNNs more easily.

\subsection{Graph Structure Attribution}
Unlike existing methods \cite{ying2019gnnexplainer, luo2020parameterized} that extract essential subgraphs via mask learning processes, we approach the graph structure attribution analysis as a link analysis problem, which is well-studied in graph research. As presented in \cref{structure_algorithm}, the attribution analysis procedure consists of only four steps. First, knowledge from model $f$ is distilled to a graph-based surrogate model $g$ via the paradigm presented in \cref{fig:knowledge_distillation}. After training $g$, we extract a learned matrix representing interactions among nodes. As this matrix is row-wise normalized, we implement a customized version of the PageRank \cite{page1999pagerank} algorithm with an initialized preference vector $\pi$ to measure node importances. After that, we visualize the output ranks $\mathcal{R_V}$ and highlight the top-$k$ nodes, where $k$ is an arbitrary parameter. If $\pi$ is uniformly distributed to all nodes in the graph, the top-$k$ nodes can be considered anchor points or influential seeds of the graph. Besides, we can personalize an instance explanation by manipulating the preference vector $\pi$, similar to the Personalized PageRank algorithm.

\begin{algorithm}[htbp]
  \caption{Graph Structure Attribution Analysis Procedure}
  \begingroup
      \raggedright
      \textbf{Input}: $f, \mathcal{G}, \pi$ \\
      \textbf{Output}: $g, \mathcal{R_V}$ \\
  \endgroup
  \begin{algorithmic}[1]
  \STATE{$g$ = distill($f$)} \COMMENT{Distill knowledge from $f$ to $g$}
  \STATE{$\mathcal{A}^*$ = extractAdjacency($g$)} \COMMENT{Extract learned interaction matrix}
  \STATE{$\mathcal{R_V}$ = PageRank($\mathcal{A}^{*T}$, $\pi$}) \COMMENT{Execute PageRank on $\mathcal{A}^*$ and $\pi$}
  \STATE{visualize($g, \mathcal{R_V}, \mathcal{A}^* $)}
  \end{algorithmic}
\label{structure_algorithm}
\end{algorithm}

To make \cref{structure_algorithm} work, we have to carefully design the student model, in which $\mathcal{A}^*$ requires two conditions: interaction representation and shared across layers. We propose two possible candidates that satisfy these conditions: GCN-LPA \cite{wang2021combining} and SGAT, a customized version of GAT \cite{velivckovic2017graph}. First, Wang et al., 2021 \cite{wang2021combining} introduced a GCN-based model that combines GCN \cite{kipf2016semi} with the label propagation algorithm (LPA). This model trains the adjacency matrix in the original GCN formula to capture feature and label influences among nodes. Second, we introduce SGAT, a modified version of GAT \cite{velivckovic2017graph}, wherein only the first layer's attention with 1-head is computed and shared across layers, inspired by \cite{xiao2019sharing}. One may think that why don't we measure node interaction using attention heads in GAT directly. For instance, Ying et al., 2019 \cite{ying2019gnnexplainer} averaged all attention heads across layers. However, we argue that this process can change the edge importance distribution.

\section{Experiments}\label{experiments}
\subsection{Experimental Setups}

\begin{table}[ht]
    \centering
    \begin{tabular}{c|c|c|c|c}
         \hline
         \thead{\textbf{Dataset}} & \thead{\textbf{\#Nodes}} & \thead{\textbf{\#Edges}} & \thead{\textbf{\#Node Features}} & \thead{\textbf{\#Classes}}\\
         \hline
         Cora & 2,708 & 10,556 & 1,433 & 7 \\
         Citeseer & 3,312 & 9,228 & 3,703 & 6\\
         Pubmed & 19,717 & 44,338 & 500 & 3\\
         \hline
         YelpChi (R-U-R)  & 45,954 & 98,630 & 32 & 2\\
         Amazon (U-P-U)  & 11,944 & 351,216 & 25 & 2 \\
         \hline
    \end{tabular}
    \caption{Dataset Information.}
    \label{tab:dataset}
\end{table}

\noindent \textbf{Datasets.} We selected two types of datasets: citation graphs and fake review detection \cite{dgldata}. We observed that node features in citation graphs were constructed from the bag-of-words model representing the most common words in documents. Therefore, finding important features for individual predictions or visualizing a global view of feature attributions is less intuitive. Instead, we focused on analyzing graph structures to provide appropriate explanations. Inversely, fake review graphs contained handcrafted features and were constructed based on mutual interactions, such as reviews from the same user having connections to each other (R-U-R) or users commenting on the same product link to each other (U-P-U). Therefore, we use these datasets to evaluate the feature attribution procedure. Detailed information on all datasets is listed in \cref{tab:dataset}.

\noindent \textbf{Evaluation Metrics.} We used seven metrics to evaluate proposed methods in different scenarios, including those listed in \cref{structure_cont}, \cref{fidel_metrics}, ROC-AUC (AUC), and Recall (R). Besides, accuracy (ACC), agreement (ARG), and $\Delta_{acc}$ were scaled to percentage. Experiments used different metrics depending on particular configurations.

\begin{equation}
    \begin{aligned}
        \textrm{Accuracy} &= \frac{1}{N} \sum_{j=1}^{N} \mathbbm{1} (\hat{y}^j = y^j) \\
        \Delta_{acc}(\mathcal{I}_i) &= \frac{1}{N} \sum_{j=1}^{N} (\mathbbm{1}(\hat{y}^j = y^j) - \mathbbm{1}(\hat{y}^j_{\backslash \{i\}} = y^j)) \\
        \textrm{Agreement} &= \frac{1}{N} \sum_{j=1}^{N} \mathbbm{1} (\hat{y}^j_s = \hat{y}^j_t) \\
        \textrm{Predictive KL} &= \frac{1}{N} \sum_{j=1}^{N} KL(\hat{p}^j_t || \hat{p}^j_s)
    \end{aligned}
    \label{fidel_metrics}
\end{equation}

\noindent \textbf{Overall Configurations.} We implemented GNN models using DGL \cite{dgldata} v0.8 and followed its example code for parameter settings. All experiments were run on an NVIDIA DGX A100 server with 4 Tesla V100 GPUs. GNN models were trained for 200 epochs most of the time without stopping using the Adam optimizer with a learning rate of 0.01 and weight decay of 5e-4. We report exceptional configurations in particular experiments. 

\subsection{Component-level Attribution Analysis}

\noindent \textbf{Settings.} We selected APPNP for the baseline model as it consists of a multi-layer perceptron (MLP) first handling features and a propagation process via the graph structure. The MLP model has one input layer with a hidden size of 64 and one output layer. All models worked with citation graphs. We measured model performance using MC and $\Delta_{acc}$. Specifically, absolute marginal contribution values of $N$ samples were averaged. To perturb node features, we replaced them with all-one vectors. Besides, the reference graph was generated by masking out all adjacent edges except node self-edges.

\begin{table}[ht]
    \centering
    \begin{tabular}{c|c|c|c|c|c}
        \hline
         & \multirow{2}{*}{\textbf{Baseline}} & \multicolumn{2}{c|}{\textbf{Feature}} & \multicolumn{2}{c}{\textbf{Structure}} \\
         \cline{3-6} 
         &                           & $\Delta_{acc}$  &   MC  & $\Delta_{acc}$ & MC \\
         \hline
        Cora & 83.1 & 68.7 & 0.48 & 11.6 & 0.20 \\
        Citeseer & 71.9 & 64.2 & 0.48 & 6.2 & 0.13 \\ 
        Pubmed & 79.3 & 61.3 & 0.54 & 6.1 & 0.22 \\
         \hline
    \end{tabular}
    \caption{The Contribution of Components in Prediction Accuracy on Citation Graphs. }
    \label{tab:different_contribution}
\end{table}

\cref{tab:different_contribution} presents the results of component attributions to the model performance on citation graphs. We can see that each component affects the model performance differently. In citation graphs, node features are essential in node classification as they contain textual information of papers encoded in one-hot vectors. Therefore, their influences are noticeably greater than the graph structures. Furthermore, the changes in model accuracy ($\Delta_{acc}$) are comparable to the accuracy of MLP models listed in \cref{tab:fidelity_res}, which verifies our approach for component attribution measurement. Surprisingly, MLP's accuracy on Pubmed is significantly larger than the feature attribution captured by $\Delta_{acc}$ and MC. We guessed that the selection of references caused this discrepancy. Despite being aware of this problem, reference selection is an arduous task requiring domain knowledge. Furthermore, citation graphs use one-hot vectors with large dimensions as nodes features, thus making this task more difficult. As these results suggested, component attribution analysis helps us find out which component is essential compared to one another by that we can invest time in data engineering or quality enhancement to improve the overall model performance.

\subsection{Fidelity of Surrogate Models}

\begin{table*}[!t]
    \centering
    \begin{tabular}{c|c|c|c|c|c|c|c|c|c|c}
        \hline 
         \multirow{2}{*}{\textbf{Modes}} & \multirow{2}{*}{\textbf{Models}} & \multicolumn{3}{c|}{\textbf{Cora}} & \multicolumn{3}{c|}{\textbf{Citeseer}} & \multicolumn{3}{c}{\textbf{Pubmed}} \\
         \cline{3-11}
         &                        & ACC & ARG & KL & ACC & ARG & KL & ACC & ARG & KL \\
         \hline
         \multirow{3}{*}{\textbf{Baseline}} 
            & GCN-LPA & 77.8 & - & - & 70.4 & - & - & 78.3 & - & - \\
            & SGAT & 80.4 & - & - & 69.7 & - & - & 77.0 & - & - \\
            & MLP & 59.8 & - & - & 59.9 & - & - & 73.3 & - & - \\
         \hline
         \multirow{3}{*}{\textbf{Teacher}} 
            & APPNP & 83.1 & 100 & 0 & 71.1 & 100 & 0 & 79.2 & 100 & 0 \\
            & GraphSage & 81.9 & 100 & 0 & 70.7 & 100 & 0 & 76.4 & 100 & 0 \\
            & GAT & 82.0 & 100 & 0 & 67.4 & 100 & 0 & 76.4 & 100 & 0 \\
         \hline 
         \multirow{3}{*}{\textbf{GCN-LPA\textsuperscript{*}}} 
            & APPNP & 83.0 & 98.8 & 0.062 & 71.2 & 97.8 & 0.023 & 78.7 & 96.4 & 0.047 \\
            & GraphSage & 81.8 & 98.6 & 0.110 & 71.7 & 98.2 & 0.073 & 76.1  & 89.6 & 0.083 \\
            & GAT & 81.9 & 87.4 & 0.204 & 67.4 & 82.1 & 0.143 & 76.7 & 90.0 & 0.073 \\
         \hline
         \multirow{3}{*}{\textbf{SGAT\textsuperscript{*}}} 
            & APPNP & 83.3 & 96.3 & 0.150 & 71.5 & 96.2 & 0.058 & 77.6 & 93.0 & 0.112 \\
            & GraphSage & 82.5 & 95.1 & 0.301 & 72.4 & 93.1 & 0.198 & 78.7 & 92.3 & 0.113 \\
            & GAT & 82.3 & 88.1 & 0.325 & 68.2 & 81.0 & 0.244 & 77.7 & 91.7 & 0.089 \\
         \hline
         \multirow{3}{*}{\textbf{MLP\textsuperscript{*}}} 
            & APPNP & 83.3 & 99.2 & 0.068 & 71.4 & 99.3 & 0.018 & 78.4 & 90.8 & 0.083 \\
            & GraphSage & 81.7 & 99.6 & 0.111 & 70.5 & 99.4 & 0.041 & 80.2 & 88.9 & 0.095 \\
            & GAT & 82.3 & 87.1 & 0.206 & 67.9 & 83.4 & 0.121 & 79.2 & 88.8 & 0.099 \\
         \hline 
    \end{tabular}
    \caption{Results of Teacher and Student Models on Citation Graphs. Students reach the accuracy of teachers and obtain high fidelity levels in all datasets. The \textbf{*} mark denotes the student models with corresponding teachers. For reference purposes,  we fill in 100\% for ARG and 0 for KL metrics of teacher results.}
    \label{tab:fidelity_res}
\end{table*}

\noindent\textbf{Settings.} We chose APPNP, GraphSage, and GAT for teacher models and selected GCN-LPA, SGAT, and MLP as students. GraphSage and GAT followed the overall configurations, while APPNP was similar to the previous experiment. We followed \cite{wang2021combining} to select parameter settings for GCN-LPA while configuring SGAT as similar to GAT and MLP as the same as APPNP's MLP. 

\cite{stanton2021does} shows that knowledge distillation sometimes causes a large divergence between the predictive distribution of the teacher and the student, wherein a good student does not imply high fidelity. In such cases, student models cannot be used as surrogate models for teacher interpretation. Therefore, we aim to answer two questions:
\begin{enumerate}
    \item Are student models as accurate as their teachers on graph data?
    \item Are student models trustworthy to use for explanations?
\end{enumerate}

We evaluated models based on three metrics accuracy, agreement, and KL. According to \cite{stanton2021does}, agreement and predictive KL scores correspond to student fidelity. Besides, the fidelity is inversely proportional to student generalization, which means if students outperform teachers, they are less trustworthy for explanations. In other words, if students can do some things that teachers cannot, explanations for students are not equivalent to those of teachers. A good student model must be as accurate as its teacher (high accuracy) and perfectly match the predictive distribution resulting in high agreement and low KL divergences. 

\cref{tab:fidelity_res} presents model results. As can be seen, all student models outperform their baselines and are comparable to teachers in the accuracy metric. When comparing them in agreement and KL metrics, students of APPNP and GraphSage have the best results. Surprisingly, GAT's students are inferior compared to these other ones. It's hard to tell what causes these results, especially with the complexity of graph data. \cite{stanton2021does} naively suggests that increasing the student training set allows it to obtain more knowledge from the teacher. We guessed that the GAT teacher generates noise in its predictions, thus making it hard for students to mimic its behaviors correctly. Besides, out of two graph-based students, GCN-LPA is superior to SGAT as the label propagation process takes advantage of label similarity among nodes in citation graphs. However, SGAT is more applicable to scenarios requiring inductive training and testing as the model computes edge weights directly from node embeddings.

\subsection{Measuring Node Feature Attributions}

\noindent\textbf{Settings.} We chose two fake review detection datasets, taken from \cite{dou2020enhancing, zhang2020gcn, rayana2015collective}, and described their top-10 features in \cref{tab:column_def}. Please refer to \cite{zhang2020gcn, rayana2015collective} for a more detailed description. In training, we applied the class-weighted technique to the GCN teacher and the oversampling method to students as datasets were highly imbalanced. Besides, we trained MLP students for 1000 epochs. For evaluation, we used three metrics: ROC-AUC (AUC), Recall (R), and agreement. 


\begin{table}[htbp]
    \centering
    \resizebox{\columnwidth}{!}{%
    \begin{tabular}{c|c|l}
        \hline
        \textbf{Dataset} & \textbf{Column} & \textbf{Description} \\
        \hline
         \multirow{10}{*}{Amazon} & MNUV   & Min. of unhelpful votes\\
         & DGP    & \#days between user first and last rating\\
         & LFS    & Length of feedback summary \\
         & MDR    & Median of rates given by the user\\
         & MXR    & Max. of rates given by the user\\
         & MNR    & Min. of rates given by the user\\
         & \#5S   & \#5-star given by the user \\
         & LU     & Length of username \\
         & \#1s   & \#1-star given by the user \\
         & STM    & Text sentiment of reviews\\
         \hline
         \multirow{10}{*}{Yelp} 
         & ISR    & If review is user’s sole review  \\
         & L      & Review length in words \\
         & Rank   & Rank order among all the reviews of product \\
         & UAvgRD & User's Avg. rating deviation \\
         & PMNR   & Product's Max. \#reviews written in a day \\
         & UMNR   & User's Max. \#reviews written in a day  \\
         & EXT    & Extremity of rating, high vote vs low vote \\
         & UBST   & Burstiness, spammers are short-term members \\
         & UWRD   & Weighted rating deviation by recency of user \\
         & PETG   & Entropy of temporal gaps of a product’s reviews \\
         \hline
    \end{tabular}
    }
    \caption{Top-10 features in Yelp and Amazon datasets. Please refer to \cite{rayana2015collective, dou2020enhancing, zhang2020gcn} for more information.}
    \label{tab:column_def}
\end{table}

\begin{figure}[htbp]
    \centering
    
    \subfloat[Amazon]{\includegraphics[width=0.97\columnwidth,trim={0.1cm 0cm 0.4cm 0.7cm},clip]{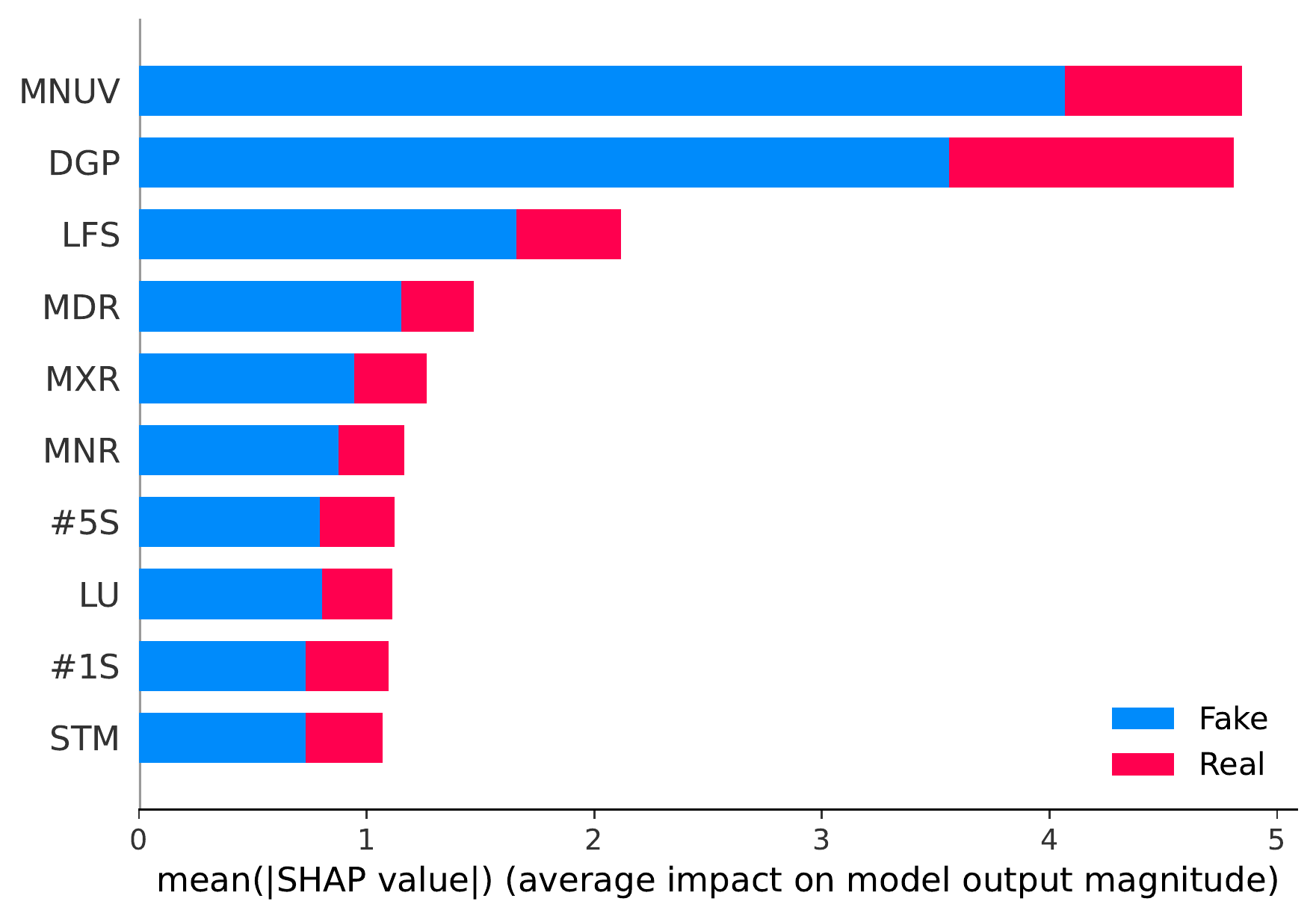}} \hfill
    \subfloat[Yelp]{\includegraphics[width=\columnwidth,trim={0.1cm 0cm 0.4cm 0.7cm},clip]{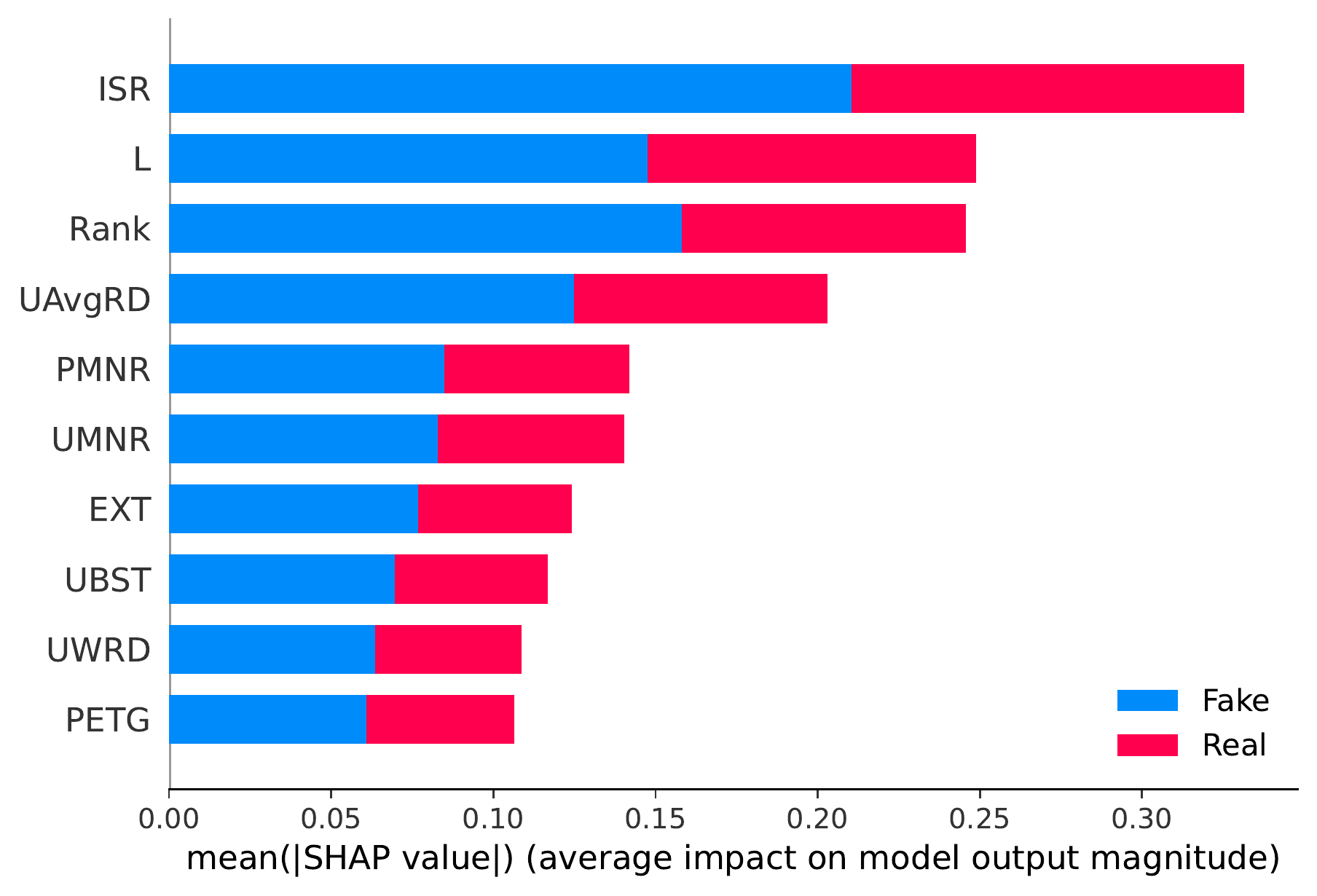}}
    \caption{Feature Attributions in the Fake Review Detection Task. We only display top-10 important features attributing to model predictions for two corresponding datasets. }
    \label{fig:yelp_amazon}
\end{figure}

Here, we aim to verify whether feature attributions can be accurately measured by the procedure presented in \cref{fig:feature_attribution}. After distilling knowledge from the GCN teacher to a simple MLP model, we executed SHAP on the student model to obtain a global understanding of feature attributions in two datasets, as presented in \cref{fig:yelp_amazon}. In the Yelp dataset, ISR (user’s sole review), L (review length), and Rank (among a product's reviews) are top-3 contributed attributes. Similarly, for the Amazon dataset, the minimum number of unhelpful votes of a user and the day gap are the two most influential factors. Without domain knowledge and a clear understanding of the problem, it's hard to judge whether attribution plots make sense. Therefore, we re-train the student model with only top-5 features for both datasets. The results, listed in \cref{tab:feature_kg}, confirm the correctness of the attribution process as the top-5 re-trained student is almost as accurate as the original student model.

\begin{table}[htbp]
    \centering
    \resizebox{\columnwidth}{!}{
    \begin{tabular}{c|c|c|c|c|c|c|c|c}
        \hline
          \multirow{2}{*}{} & \multicolumn{2}{c|}{\thead{ \textbf{Teacher} \\ \textbf{(GCN)} }} 
                             & \multicolumn{3}{c|}{\thead{\textbf{MLP Student} \\ \textbf{(All features)} }}
                            & \multicolumn{3}{c}{\thead{\textbf{MLP Student} \\ \textbf{(Top-5 features)}}}\\
          \cline{2-9} 
          
                & AUC & R       & AUC & R & AGR & AUC & R & AGR  \\
          \hline 
         Yelp   & 80.9 & 78.2     & 74.5 & 73.5 & 85.7    & 74.7 & 73.7 & 85.5 \\
         Amazon & 81.4 & 81.5     & 88.9 & 84.2 & 83.3    & 88.3 & 84.8 & 82.3 \\
         \hline
    \end{tabular}}
    \caption{Results of the GCN Teacher and MLP Students on Yelp and Amazon Datasets. The MLP student trained with top-5 features is comparable to one trained with all features.}
    \label{tab:feature_kg}
\end{table}

\subsection{Examining Graph Structure Attributions}

To increase trust in the explainer, we implemented a visualization tool that supports interactive explanations of the graph structure. The software provides a global view of graph embedding and multi-level visualization for local explanations. Each view supports edge filtering and highlighting features for nodes and edges to provide a clear understanding of internal model logic. This section presents \cref{structure_algorithm}'s results on the Cora graph with APPNP teacher and SGAT student using this software. 

\begin{figure}[ht]
    \centering
    \includegraphics[width=\columnwidth, trim={13.7cm 0.7cm 14.7cm 0.1cm},clip]{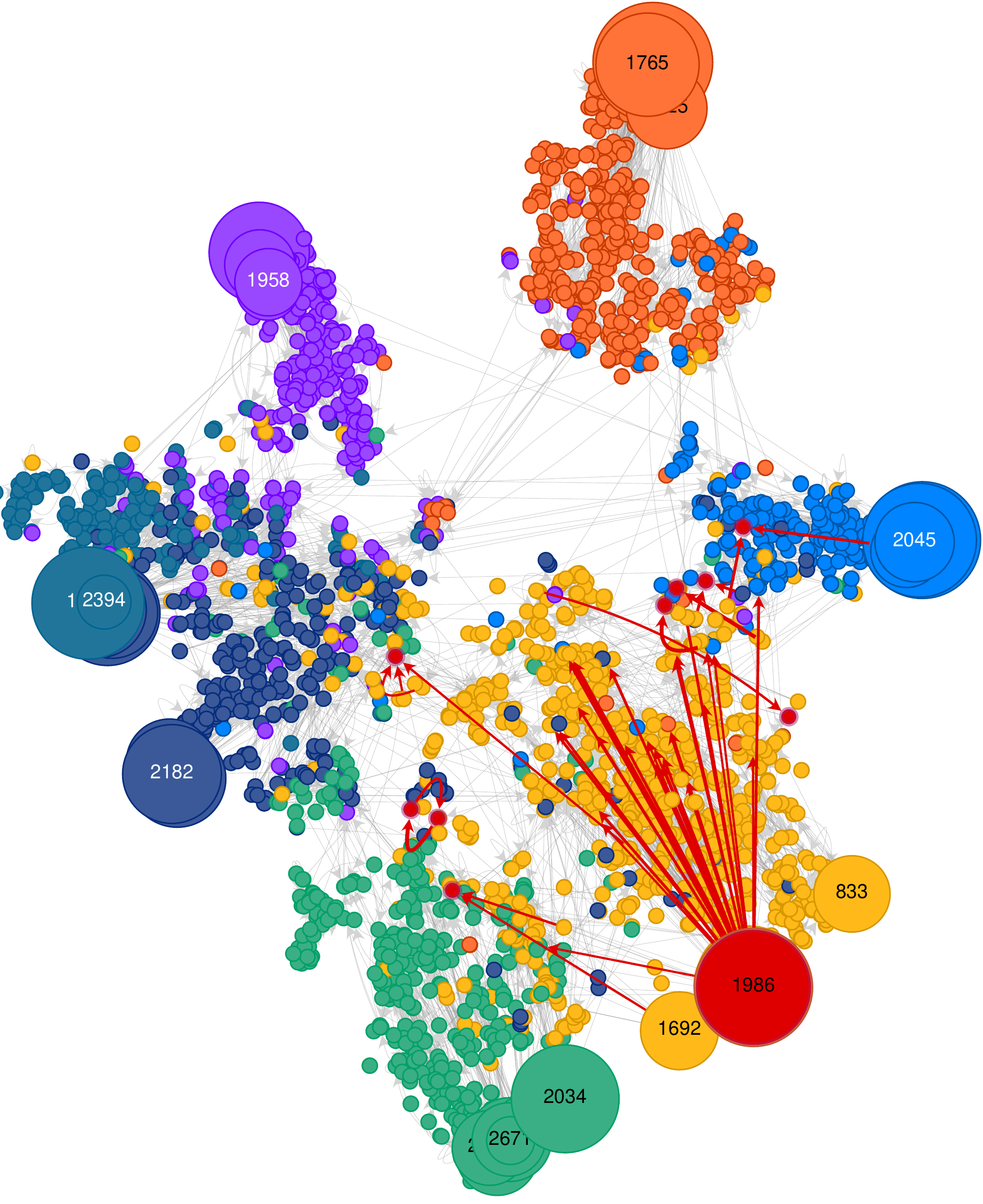}
    \caption{Global Summary Visualization of Cora Graph based on APPNP teacher and SGAT Student Model. We filter out edges with a probability smaller than 0.3 and highlight top-50 nodes (large nodes) ordered by ranks. Red nodes are selected ones, while red edges show the influential flows from/to them. Nodes in different colors are in different clusters.}
    \label{fig:global_view}
\end{figure}

As presented in \cref{fig:global_view}, nodes in the Cora graph are displayed in clusters corresponding to its embedding vectors, projected into the 2-D dimension using T-SNE. From this global view, we can detect cluster boundaries and specify influential factors on a high level. By selecting a node, we can see how it impacts on or is affected by its neighbors. Furthermore, the global view provides an overall understanding of how neighbors impact nodes located at the boundary lines in different clusters. We also observed that important nodes usually have a high degree connected to numerous nodes within their clusters and are located at extreme corners. However, an influential node in the global view only means that it connects to numerous neighbors and sometimes does not have a decisive impact on them. For instance, a top-cited paper has many links to papers across domains, but downstream nodes' classes depend on their locally connected neighbors.

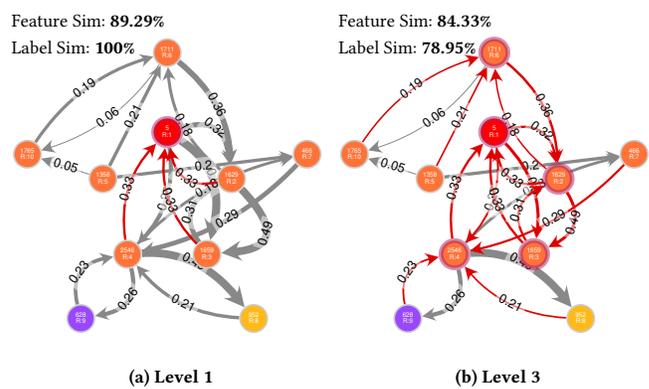
\begin{figure}[ht]
    \centering
    \subfloat[Level 1\label{local_a}]{
        \includegraphics[width=0.485\columnwidth, trim={15cm 3cm 13cm 0cm},clip]{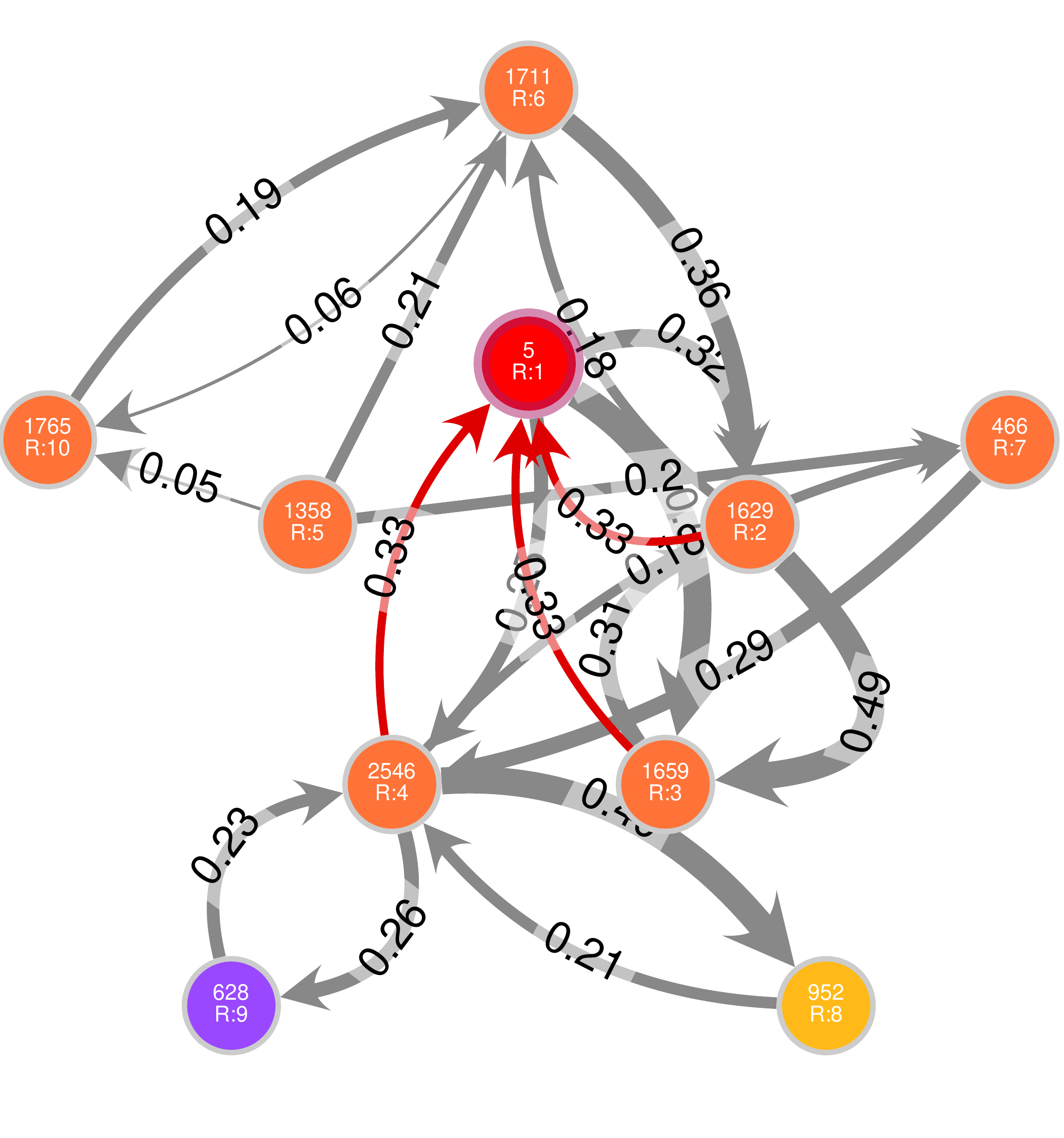} 
        \begin{picture}(0,0)
            \put(-120,125){\footnotesize Feature Sim: \textbf{89.29\%}}
            \put(-120,115){\footnotesize Label Sim: \textbf{100\%}}
        \end{picture}
    } 
    \subfloat[Level 3\label{local_b}]{
        \includegraphics[width=0.485\columnwidth, trim={15cm 3cm 13cm 0cm},clip]{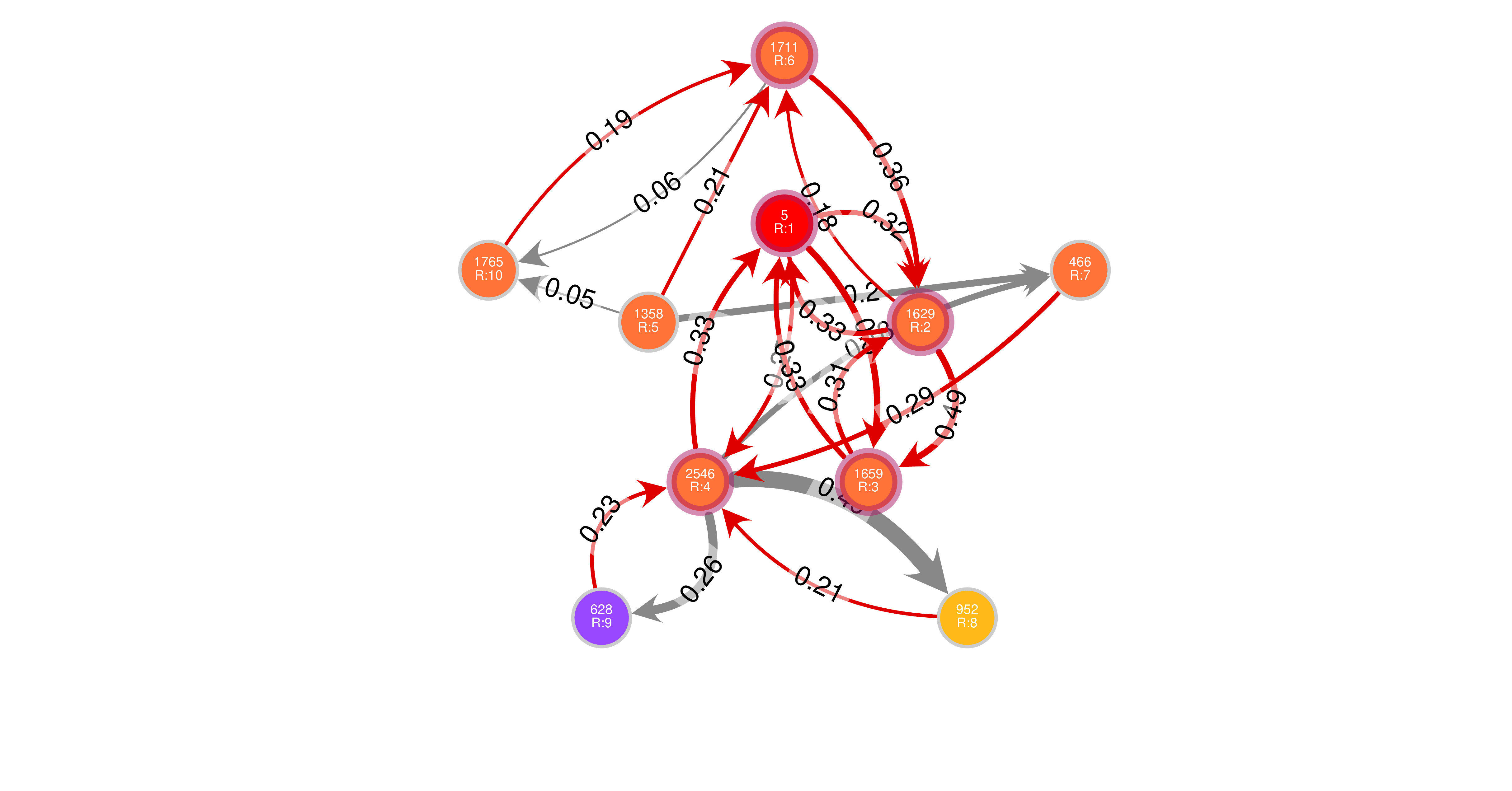} 
        \begin{picture}(0,0)
            \put(-120,125){\footnotesize Feature Sim: \textbf{84.33\%}}
            \put(-120,115){\footnotesize Label Sim: \textbf{78.95\%}}
        \end{picture}
    }
    \caption{Local Explanations of Node ID 5 (Red Node). Each level shows the contributions of corresponding k\textsuperscript{th}-hop neighbors. Edge weights are normalized probabilities. }
    \label{fig:local_id5}
\end{figure}

\begin{figure}[ht]
    \centering
    \subfloat[Hierarchy]{
        \includegraphics[width=0.49\columnwidth,trim={13.5cm 1cm 11cm 0cm},clip]{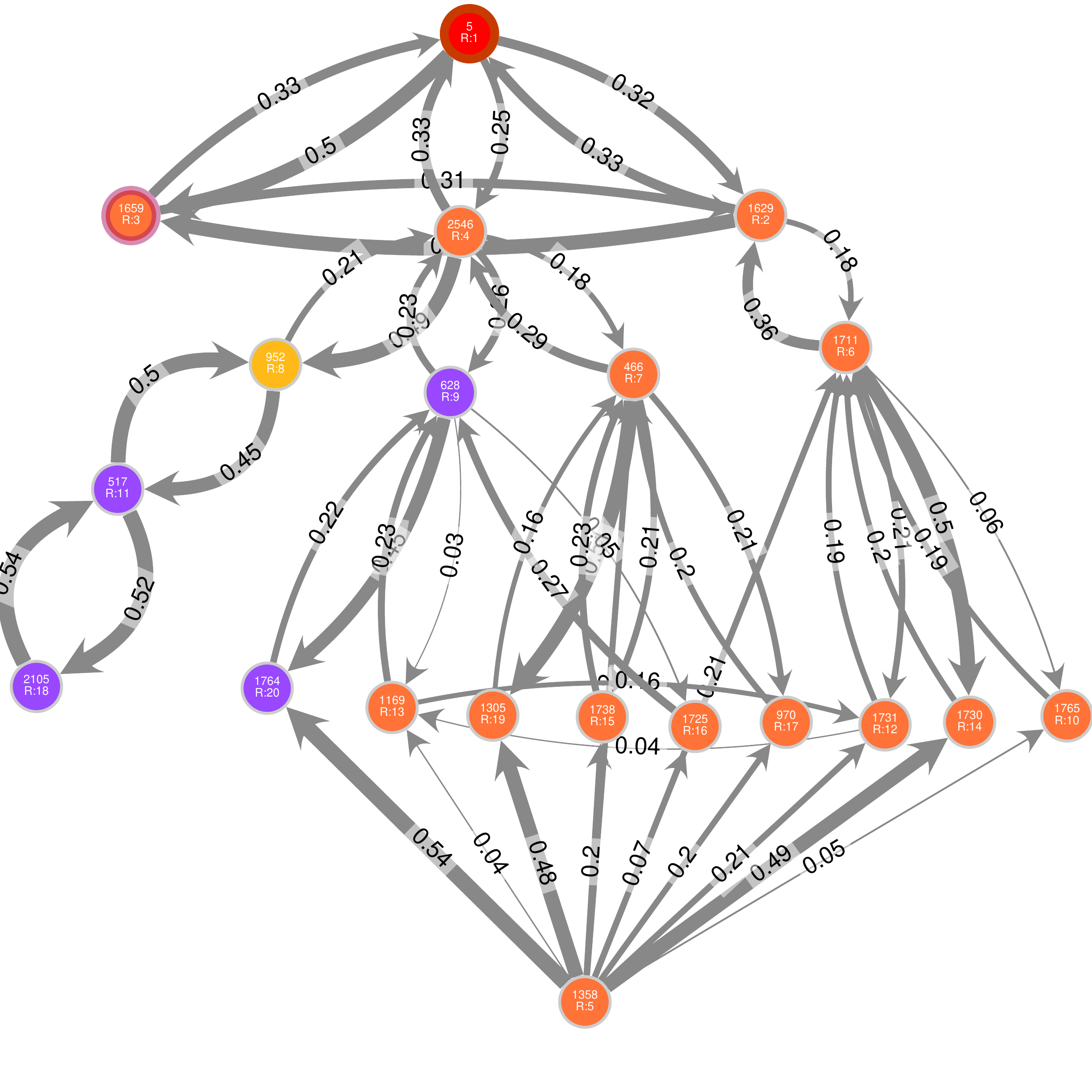}
    } 
    \subfloat[Force]{\includegraphics[width=0.5\columnwidth,trim={14cm 1.5cm 12.5cm 1.2cm},clip]{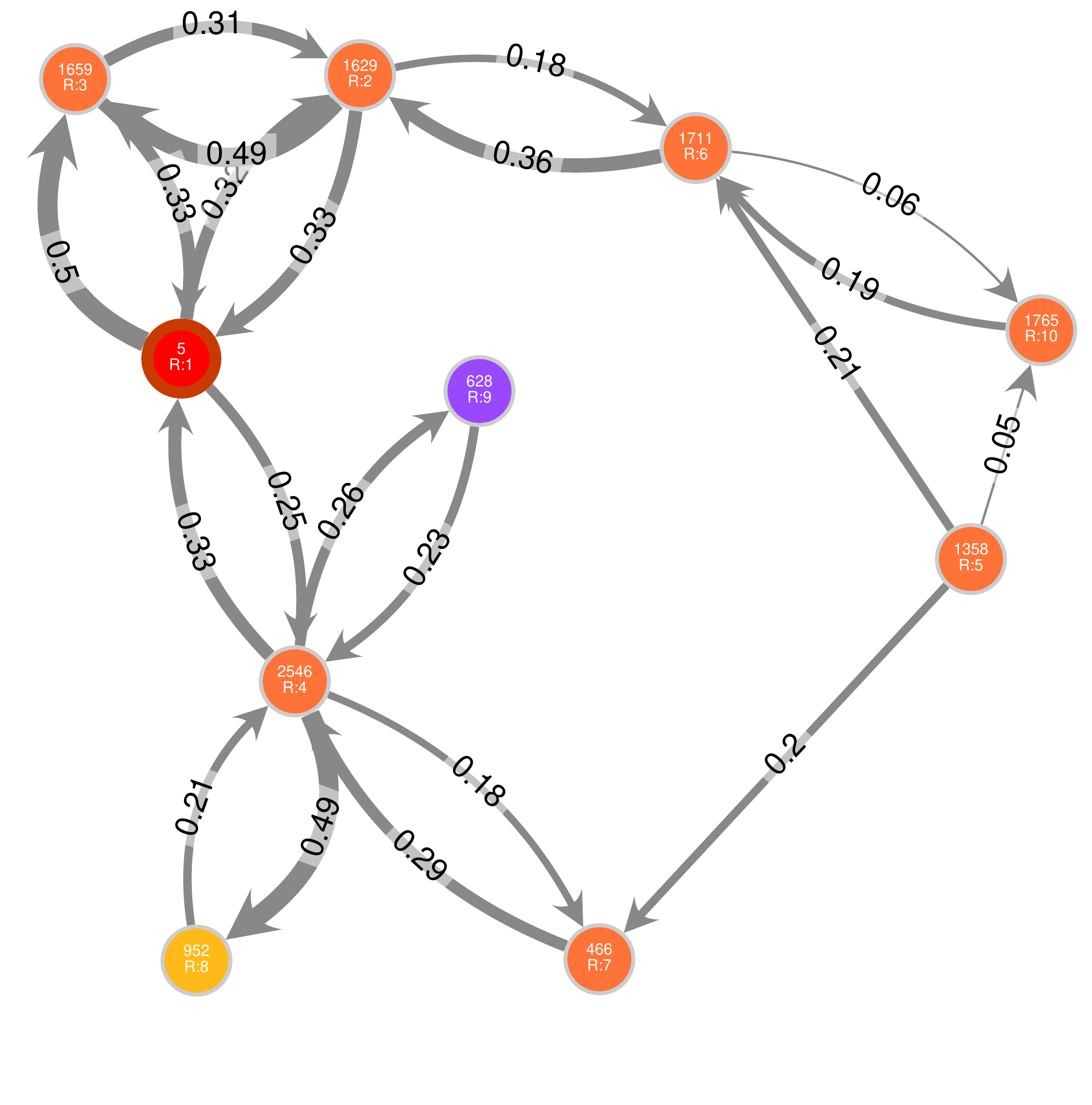}}
    \caption{Different Graph Layouts for a Local Explanation of Node 5 (Red Node) in Cora Graph.}
    \label{fig:graph_layout}
\end{figure}

Similarly, \cref{fig:local_id5} depicts multiple levels of local explanations for an example node. In local explanation, our software provides additional features such as graph layout selection, k\textsuperscript{th}-hop highlighting, edge filtering, feature similarity, and label similarity \cite{liu2020alleviating} to support various requirements. First, we can adjust the number of visible neighbors in one view ordered by the Personalized PageRank algorithm rooted as the target node. With this feature, we present influential flows from neighbors to the target node at different levels. For simplicity, we only show two levels in \cref{fig:local_id5}. Secondly, feature and label similarities, shown in the top left corner of this figure, present the purity of the community rooted at a target node, which provides another perspective of how the prediction was made. Finally, multi-layout graph visualization is crucial as users may prefer different views of the same output, as shown in \cref{fig:graph_layout}. 

\subsection{Examining Online Knowledge Distillation}
\noindent\textbf{Settings.} We adopted the training paradigm from \cite{zhu2018knowledge}. For simplicity, we selected only the best teacher APPNP with two students, GCN-LPA and MLP. All models were trained on citation graphs using transductive settings.

\begin{table}[ht]
    \centering
    \begin{tabular}{c|c|c|c|c|c|c|c}
        \hline
         & \thead{\textbf{Teacher} \\ \textbf{APPNP}} & \multicolumn{3}{c|}{\thead{\textbf{Student} \\ \textbf{GCN-LPA} }}  & \multicolumn{3}{c}{\thead{\textbf{Student} \\ \textbf{MLP} }} \\
         \cline{2-8}
         & ACC & ACC & ARG & KL & ACC & ARG & KL \\ 
         \hline
         Cora & 83.5 & 81.7 & 88.8 & 0.159 & 75.0 & 76.1 & 1.568\\
         Citeseer & 72.4 & 69.1 & 83.5 & 0.113 & 69.9 & 73.8 & 0.471 \\
         Pubmed & 78.9 & 76.4 & 91.5 & 0.086 & 76.2 & 84.1 & 0.386\\
         \hline
    \end{tabular}
    \caption{Performance of Models in the Online Knowledge Distillation Task. Here, agreement scores are lower than the scores obtained by the offline mode.}
    \label{tab:online_kg}
\end{table}

In this experiment, we aim to verify the performance of online knowledge distillation on graph data since it enables further extensibility and scalability. We observed that the ensemble paradigm \cite{zhu2018knowledge} requires models to be homogeneous to operate efficiently, especially on graph data. Specifically, when we trained only the teacher with graph-based students, both worked perfectly. However, when an MLP student was included in the ensemble, it achieved inferior performance and sometimes degraded the others' accuracy. Therefore, we had to apply a few modifications to obtain acceptable results for MLP models, such as multiplying its loss by a factor or computing KL Loss between its output probabilities to all models. 

As shown in \cref{tab:online_kg}, all models obtain low agreement scores compared with their teachers. This observation suggested that even though online knowledge distillation can help student models achieve teacher accuracy, its fidelity level is significantly lower than the offline mode. Therefore, this trade-off should be carefully considered in real-world scenarios to balance system performance and accuracy benefits.

\section{Discussion}\label{discussion}

Recently, explainable AI (XAI) has been an active research area as explanation is now a crucial part of advanced AI systems (software 2.0 \cite{dilhara2021understanding}). We argue that XAI for GNN is only at the early stage, and there is plenty of room for improvement. When starting this work, we observed that most current methods linked to \cite{ying2019gnnexplainer} proposed to train an explainer using masking methods on synthesized datasets with handcrafted ground-truth information. However, explanation ground truths are difficult to obtain in real-world scenarios. Furthermore, explanations are selected and don't refer to statistical relationships due to biases \cite{miller2019explanation}. Hence one explanation does not fit all people. Besides, an interactive visualization tool, which provides multiple meaningful information, is needed to enhance consensus between users and the explanation system.

Many feature attribution methods have been proposed lately, such as \cite{lundberg2017unified,tsang2020does}. We thought about how to use them for GNN explanations. However, it is not easy to apply these methods to GNN models directly due to computations in the graph structure. We found that knowledge distillation works perfectly with GNNs, wherein we have various options for student models. Therefore, the idea is to transform the original GNN models into interpretable versions. A similar approach is applied to the graph structure attribution analysis with numerous efficient methods such as PageRank \cite{page1999pagerank}.

Our work has a few limitations. First, we haven't elaborated on fine-grained interactions among features and the graph structure. Even though we found that node features are crucial to the prediction performance, how the model performance got enhanced when including the graph structure was not analyzed. Besides, we only tackled the node classification problem while untouched the other problems such as graph classification or link prediction. These challenges will be potential research topics.

\section{Conclusion}\label{conclusion_part}
In this paper, we proposed a multi-level GNN explanation framework. 
We observed that graph data is a composition of entities regarded as nodes and relationships between them, which form the graph structure. Therefore, we considered GNN as a multimodal learning process of these components. Next, we decomposed the explanation problem into multiple sub-problems, formulating a hierarchical structure. At the top level, the framework specified the contribution of each component to GNN execution and predictions. In contrast, fine-grained levels focused on feature attribution and graph structure attribution analysis based on knowledge distillation. Furthermore, we also aimed to personalize explanations by providing customized outputs for different preferences. To confirm our hypotheses, we conducted extensive experiments and showed that the proposed approach is promising and extendable.


\bibliographystyle{ACM-Reference-Format}
\bibliography{bibliography}

\end{document}